\documentclass[journal]{IEEEtran}
\usepackage{amsmath,graphicx,amssymb,bm}
\usepackage{subfig}
\usepackage[T1]{fontenc}

\usepackage{color}

\begin{document}
%
\title{Precision Learning: Reconstruction Filter Kernel Discretization}
\author{C.~Syben, B.~Stimpel, K.~Breininger, T.~Würfl, R.~Fahrig, A.~Dörfler and A.~Maier%
\thanks{This work has been supported by the project P3-Stroke, an EIT
				Health innovation project. EIT Health is supported by EIT, a body
				of the European Union.}%
\thanks{C. Syben, B. Stimpel, K.Breininger, T. W\"urfl, R.~Fahrig and A. Maier are with Friedrich-Alexander-Universit\"at Erlangen-N\"urnberg, Pattern Recognition Lab, Erlangen, Germany.}
\thanks{ C. Syben, B. Stimpel, and A. D\"orfler are also with Friedrich-Alexander-Universit\"at Erlangen-N\"urnberg, Department of Neuroradiology, Erlangen, Germany.}
			}

\maketitle
\pagestyle{empty}
\thispagestyle{empty}
\begin{abstract}
In this paper, we present substantial evidence that a deep neural network will intrinsically learn the appropriate way to discretize the ideal continuous reconstruction filter. Currently, the Ram-Lak filter or heuristic filters which impose different noise assumptions are used for filtered back-projection. All of these, however, inhibit a fully data-driven reconstruction neural network learning approach. In addition, the heuristic filters are not chosen in an optimal sense. To tackle this issue, we propose a formulation to directly learn the reconstruction filter. 
The filter is initialized with a Ramp filter as a strong pre-training and learned in frequency domain. We compare the learned filter with the Ram-Lak and the Ramp filter on a numerical phantom as well as on a real CT dataset.
The results show that the network properly discretizes the continuous Ramp filter and converges towards the Ram-Lak solution. In our view these observations are interesting to gain a better understanding of deep learning techniques and traditional analytic techniques such as Wiener filtering and discretization theory. Furthermore, this will allow fully trainable data-driven reconstruction deep learning approaches.

\end{abstract}
%
%
\section{Introduction}
\label{sec:intro}
Recently deep learning has shown promising results in the field of Computed Tomography (CT) reconstruction. In his perspective article, Wang \cite{Wang2016} states that a reconstruction pipeline implemented as a deep neural network allows to access the capability of learning-based reconstruction. Wang identifies the data-driven knowledge-enhancing abilities as the strength of deep learning-based reconstruction. Würfl et al. \cite{Wuerfl2016} have proposed an implementation of the filtered back-projection algorithm (FBP) as a neural network. Similar to iterative reconstruction algorithms, their proposed implementation avoids explicitly storing the system matrix, which would render the algorithm infeasible.
The approach utilized the data-driven capability by learning the compensation weights in case of limited-angle tomography. Hammernik et al. \cite{Hammernik2017} proposed a two-level deep learning architecture to compensate for additional streak artifacts in the limited-angle tomography case. They showed that their approach allows for a joint optimization without any heuristic parameter tuning. In both approaches the necessary filtering to perform a FBP is done with a fixed layer using an analytical discretization. 

\begin{figure}[t]	
	\centering
	\subfloat{\includegraphics[width=0.45\textwidth]{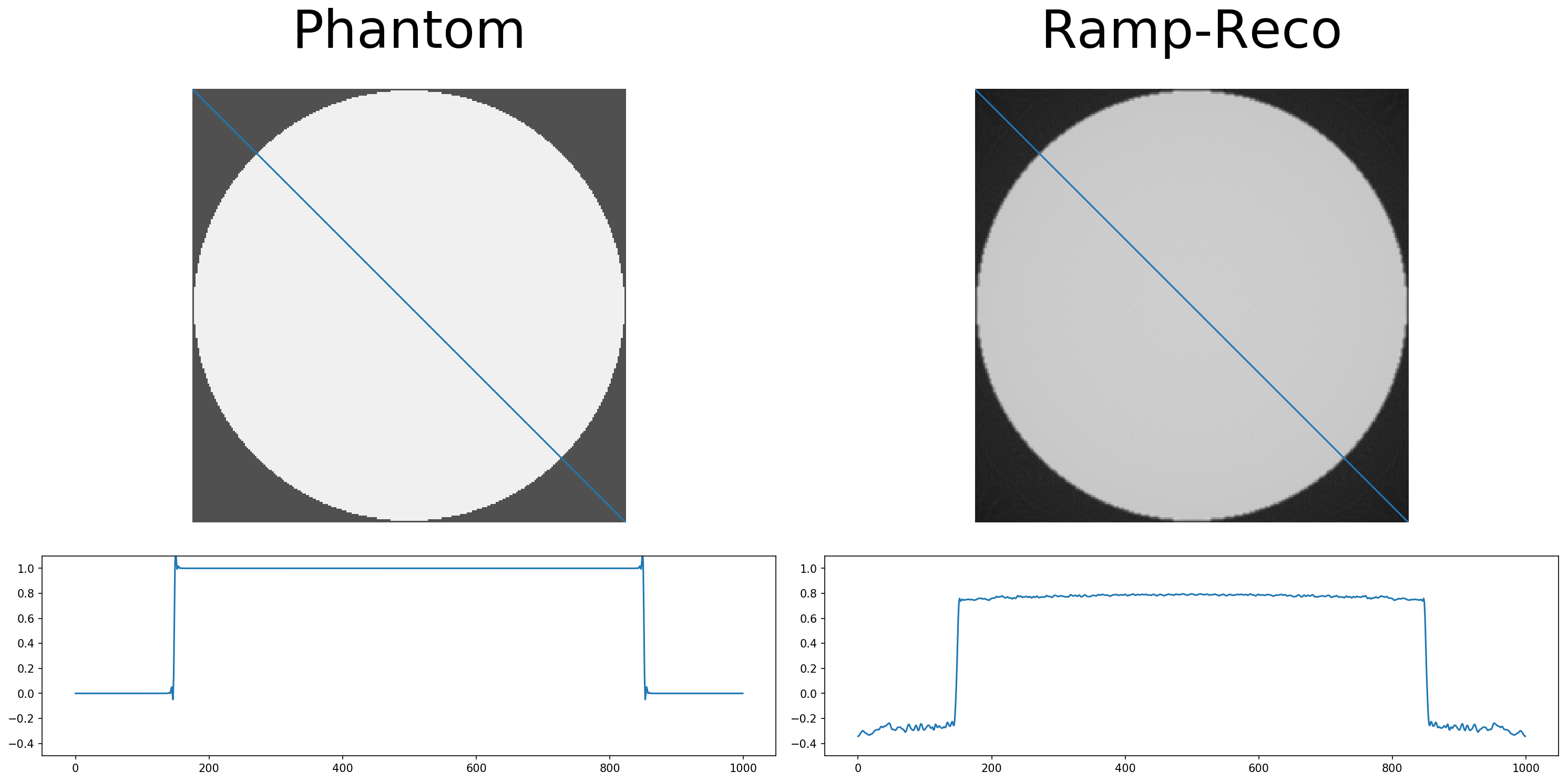}}
	\caption{Line profile through a 2D circle phantom. Ground truth (left) and FBP result based on Ramp filter (right).}
	\label{ramp_error}
\end{figure}

The ideal filter for FBP can be derived using analytic reconstruction theory. Assuming an infinite number of projections and infinitely small detector pixels, it takes the form of the absolute value function in Fourier domain, commonly referred to as Ramp filter. In practice, however, Radon inversion has to be performed using a finite number of projections. This introduces discretization errors which need to be handled by replacing the Ramp filter with an appropriate discrete version. The occurring artifacts are commonly called cupping and dc shift artifacts \cite{kak}, and an example of both is shown in Figure \ref{ramp_error}.
The analytically derived discrete version of the Ramp filter is the well known Ram-Lak filter introduced by Ramachandran and Lakshminarayan \cite{RamLak}. Both, the ideal Ramp as well as the Ram-Lak filter are sensitive to noise. Thus, many different filters have been proposed which impose different noise assumptions. The most well known example is the Shepp-Logan filter \cite{SheppLogan}, 
which incorporates a smoothing filter. However, while the choice of filter clearly matters, none of these heuristic filters are chosen in an optimal sense. CT vendors typically have specialized departments that aim at designing optimal filters for their users. The actual filter configurations are often deemed as company secrets. This shows the demand for specialized filters and raises the question whether data-driven methods can be used to deduce optimal filters.  

Pelt et al. \cite{Pelt2014} proposed a method to learn such a data-dependent filter for the FPB algorithm. Their approach learns a filter approximation to increase reconstruction accuracy in specific cases. They optimize the filter with respect to the minimal error between the input data and the projection of the FBP reconstructed input. Pelt et al. use an efficient formulation and exponential binning to handle the size of the system matrix. In order to do this, handling the discretization of the Ramp filter with respect to the details around the zero frequency as well as the boundaries is necessary.

We propose a formulation to learn a discrete optimal reconstruction filter directly in a deep learning context. We provide substantial evidence that a deep neural network will intrinsically learn the appropriate way to discretize the continuous filter. Furthermore, our proposed formulation leads to a straightforward implementation of the optimization using deep learning frameworks that automatically compute the gradient of the cost function using back-propagation.

\section{Method \& Materials}
First, we describe the filtered back-projection in continuous and discrete form for the parallel-beam geometry and introduce the theoretical filter kernel. Afterwards, the optimization problem and the related gradient to learn the filter kernel are derived. In the last Sections we describe our experiments and discuss them.
\subsection{Filtered Back-Projection}
\label{fbp}
The filtered back-projection (FBP) algorithm is an efficient solution to the reconstruction problem. First the projection data $p(s,\theta)$ are filtered by a convolution with the filter kernel
\begin{eqnarray}
q(s,\theta) &= h(s) * p(s,\theta)\label{filtering} \enspace, \\
h(s) &= \int |\omega|e^{2\pi\omega s} d\omega \label{filter} \enspace,
\end{eqnarray}
subsequently, the filtered projection data $q(s,\theta)$ are back-projected to obtain the reconstruction result $f(x,y)$ with
\begin{equation}
f(x,y) = \int_{0}^{\pi} q(s,\theta)|_{s = x\cdot\cos\theta+y\cdot\sin\theta}d\theta \label{backprojection}\enspace.
\end{equation}
In practice, we need a discrete description of the FBP presented in Eq. (\ref{filtering}-\ref{backprojection}). The discrete reconstruction problem can be expressed by
\begin{equation}
\bm A \bm x = \bm p  \enspace ,
\end{equation}
where $\bm A\in \mathbb{R}^{N\times M\cdot P}$ is the system matrix, $\bm x\in \mathbb{R}^{N}$ is the volume and $\bm p\in \mathbb{R}^{M\cdot P}$ are the projections.
Since the discrete representation of the Radon transform $\bm A$ is a tall matrix, it has no inverse. This means every reconstruction formula is, therefore, a unique pseudo inverse of $\bm A$,
\begin{equation}
\bm x = \underbrace{\bm A^{\top}}_{\text{Back-projection (Eq. \ref{backprojection})}} \underbrace{(\bm A\bm A^{\top})^{-1}}_{\text{Filter (Eq.\ref{filtering})}} \bm p \enspace.
\end{equation}

\subsection{Learning the Filter Kernel}
\label{learn_filter}
From the convolution theorem it is clear that a convolution in spatial domain is equal to a multiplication in the frequency space. Therefore, we can reformulate the filter $(\bm A\bm A^\top)^{-1}$ in form of a diagonal matrix in Fourier domain
\begin{equation}
\bm x = \bm A^{\top} \bm F^{\bm H}\bm K\bm F\bm p 
\label{nn_formulation} \enspace,
\end{equation}
where $\bm F\in\mathbb{R}^{N\times N}$ and $\bm F^{\bm H}\in\mathbb{R}^{N\times N}$ represents the Fourier and inverse Fourier transform, respectively. $\bm K\in\mathbb{R}^{N\times N}$ is a diagonal matrix representing the filter in frequency domain.
To learn the filter matrix $\bm K$, which is a discrete approximation of the Ramp filter, we use Eq. \ref{nn_formulation} to formulate a objective function $f(\bm K)$ as a least-square minimization problem:
\begin{equation}
f(\bm K) = \frac{1}{2} \lVert \bm A^\top\bm F^{\bm H}\bm K\bm F\bm p - \bm x \rVert_2^2 \enspace.
\label{objective_function}
\end{equation}
The gradient to our objective function $f(\bm K)$ in Eq. \ref{objective_function} with respect to $\bm K$ is 
\begin{equation}
\frac{\partial f(\bm K)}{\partial\bm K} = \underbrace{\bm F\bm A\underbrace{(\bm A^T\bm F^{\bm H}\bm K\bm F\bm p - \bm x)}_{\text{Error}}}_{\text{Back-propagation}}(\underbrace{\bm F\bm p}_{l-1})^\top \enspace.
\label{back_prop}
\end{equation}
Note that this analytical gradient also has an interpretation in a neural network learning context.
To describe Eq. \ref{back_prop} using the terms of back-propagation: 
Eq. \ref{nn_formulation} can be regarded as a network with input $\bm p$ and layers $\bm F$, $\bm K$, $\bm F^H$ and $\bm A^T$ with the identity as activation function between layers. $\bm K$ is the only layer containing trainable weights. Then, $\hat{\bm x} = \bm A^T\bm F^{\bm H}\bm K\bm F\bm p$ is the forward pass through the network following the considerations by Würfl et al. \cite{Wuerfl2016}.
The gradient of the error function (Eq. \ref{objective_function}) with respect to $\bm K$ is computed by multiplying two factors: 1) the partial derivative of the error function with respect to the output of the layer and 2) the transpose of the output of the previous layer ($l-1$). The output of $l-1$ is readily described by $\bm F\bm p$. Using the recursive formulation of back-propagation, we yield ${\bm F\bm A(\hat{\bm x} - \bm x)}$, with $(\hat{\bm x} - \bm x)$ being the derivative of the error function.
We consider these observations as interesting, as this gradient would be computed automatically in a deep learning framework such as TensorFlow.
\color{black}

\subsection{Experiments}
We implemented the cost function and the analytically derived gradient in CONRAD \cite{Maier.2013} and used stochastic gradient descent to learn the filter. Unmatched projectors were used for the reconstruction (pixel driven) and the forward projections (ray driven). 0
The filter matrix $\bm K$ is initialized with a slightly modified frequency domain representation of the ideal Ramp filter. By doubling the width of the zero-valued part of the ramp we emphasize the cupping artifacts. This serves exclusively to show the learning capability. 
For the training we use 10 numerical disc phantoms with increasing radii rendered on a  512$\times$512 pixel grid. The learned filter is evaluated on a 512$\times$512 slice of a real CT dataset showing the head of a pig acquired at Stanford University. For evaluation we compare the filtered back-projection using the modified Ramp Filter, the Ram-Lak and the learned filter, in the following referred to as Ramp-reco, Ram-Lak-reco and Learned-reco, respectively.
The quantitative evaluation is done on the absolute difference between the ground truth (GT) and the respective reconstruction result.
These difference images are evaluated using the mean, minimum and maximum difference as well as the standard deviation (std. dev.). Note that we use the original reconstruction as the ground truth for the pig experiment.

\section{Results}
\begin{figure*}
	\centering
	\includegraphics[width=\linewidth]{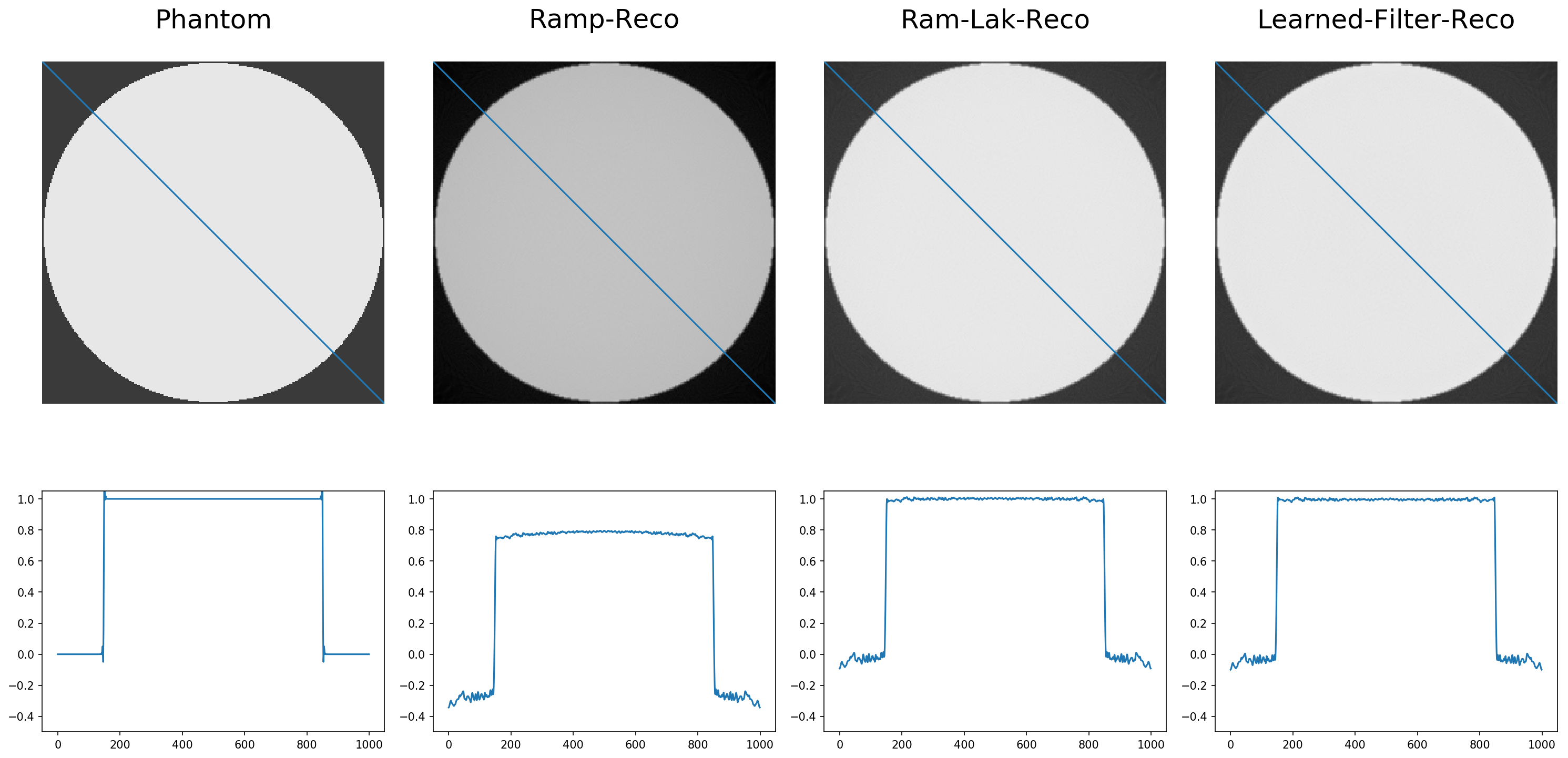}		
\caption{Reconstructions and respective line profile plots of numeric circle phantom using different filters. All images have the same window/level.}
\label{circle_results}	
\end{figure*}

\begin{figure}[htb]	
	\centering
	\includegraphics[width=\linewidth]{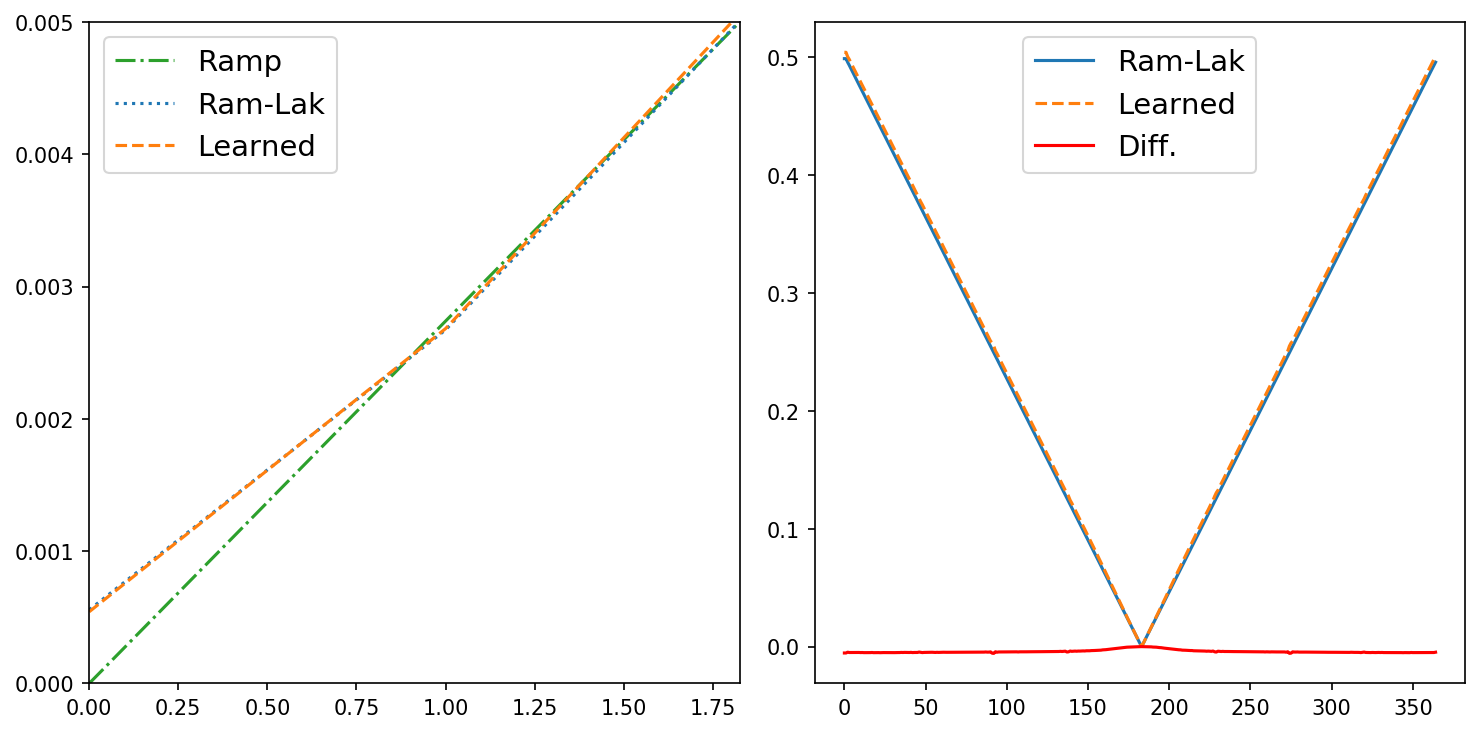}	
	\caption{Zero frequency behavior of the Ramp, Ram-Lak and learned filter.}	
		\label{filter_diff}
\end{figure}

\begin{figure}[t]	
	\centering
	\includegraphics[width=\linewidth]{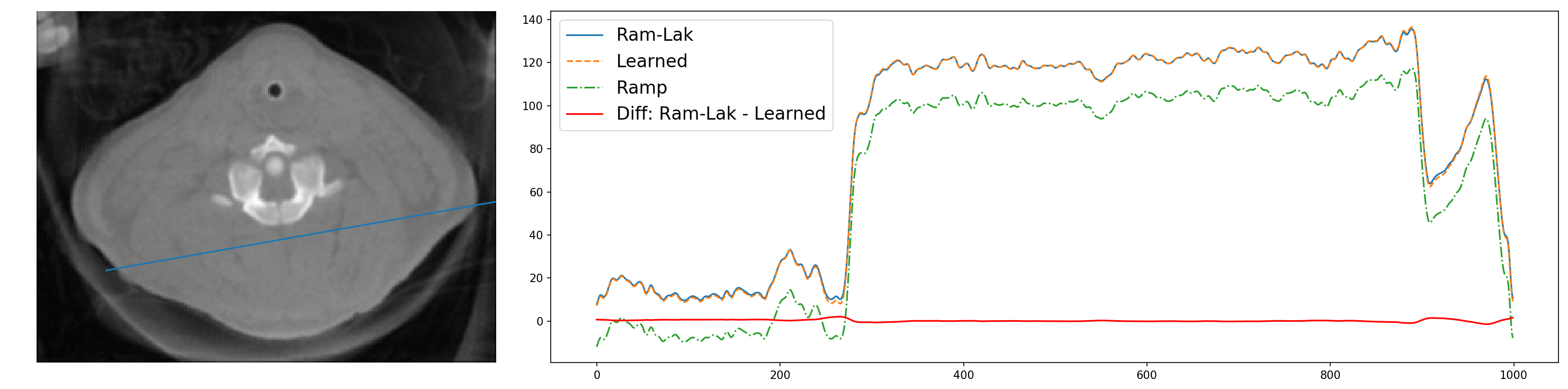}
	\caption{Reconstruction of a pig dataset filtered with Ramp, Ram-Lak and learned filter. The difference between the Ram-Lak and the learned filter is plotted in red.}	
	\label{pig_results}
\end{figure}

\subsection{Qualitative Results}
Fig. \ref{circle_results} shows the results of the FBP on a numerical disc phantom, which was used for the training. Additionally, the results of Ramp-Reco , the Ram-Lak-Reco and the Learned-Filter-Reco and their respective line profiles are presented. The Ramp-Reco leads to cupping artifacts in the homogeneous area of the disc that can be observed in Fig. \ref{circle_results} as well as in the respective line profile. The line profile for the Ram-Lak-Reco illustrates that the homogeneous area of the disc is reconstructed properly. The Learned-Filter-Reco reconstructs the homogeneous area properly as well. Comparing the line profiles of the Ram-Lak-Reco with the line profile from the Learned-Reco, the Ram-Lak-Reco shows nearly a straight line while the line profile of the Learned-Reco still shows small deviations from the ideal line.
Fig.\ref{filter_diff} show the differences of the Ramp, Ram-Lak and learned filter around the zero frequency and on the full spectrum.
In Fig. \ref{pig_results} the reference reconstruction is shown and the respective line-profiles from the different reconstructions using the Ramp, Ram-Lak and learned Filter are plotted. 
Both line profiles, aligned over the same homogeneous area of the pig, show a similar behavior and no cupping.

\subsection{Quantitative Results}
In Tab. \ref{tab_circle_diff_error} the measurements of the absolute difference between the GT and the respective reconstructions are shown. 
The absolute mean error of the Ramp-Reco is at $23.5\%$, while the absolute mean error of the Learned-Filter-Reco is at $2.3\%$. The lowest absolute mean error can be observed by the Ram-Lak-Reco. The standard deviation as well as the maximum value of the Ramp-Reco are higher compared to the other two reconstructions. The Ram-Lak-Reco and the Learned-Filter-Reco show a similar standard deviation of the absolute difference as well as the absolute maximum error.
In Tab. \ref{tab_pig_diff_error} the absolute error measurements between the GT and the respective reconstructions are presented. Both the Ram-Lak-Reco as well as the Learned-Filter-Reco exhibit a similar absolute mean error. Also the standard deviation of the absolute error as well as the absolute maximum error are closely together.
\begin{table}[t]
	\caption{Measurements of the absolute difference between GT and respective circle reconstruction (in percent).}
	\label{tab_circle_diff_error}
	\begin{tabular*}{\linewidth}{lcccc}
		\hline
						& mean 	& std. dev. & min  		& max  	\\ \hline
		Ramp-reco 		& 0.235	& 0.07		& 0.001 	& 0.596 \\ \hline
		Ram-Lak-reco 	& 0.01	& 0.031		& 0 		& 0.41	\\ \hline
		Learned-reco	& 0.023 & 0.03		& 6.76E-09 	& 0.409	\\ \hline	
	\end{tabular*}
\end{table}
\begin{table}[t]
	\caption{Measurements of the absolute difference between GT and respective pig reconstruction in Hounsfield units (HU).}
	\label{tab_pig_diff_error}
	\begin{tabular*}{\linewidth}{lcccc}
		\hline
						& mean 		& std. dev.	& min		& max  \\ \hline
		Ram-Lak-reco 	& 66.99		& 61.401	& 6.10E-5	& 1634.82 \\ \hline
		Learned-reco	& 83.53		& 68.06 	& 8.39E-5	& 1685.70 \\ \hline	
	\end{tabular*}
\end{table}
\section{Discussion}
The evaluation of the numerical disc phantom shows that homogeneous areas can be reconstructed properly without cupping artifacts when using the learned filter. However, the line profile reveals that the result is not as good as with the Ram-Lak filter. A possible explanation for this difference is our way of implementing the optimization directly and configuring the stochastic gradient parameters heuristically. Also note that our current training is performed with only 20 epochs and our data set consists only of 10 cylinders of different diameter. This is only a coarse approximation of the ideal training set. We assume that using a basis that spans the entire domain of $\bm x$ will do much better for this job. This will be subject of future experiments. Still, we consider 10 training samples as a good start for estimating such a complex relation that generalizes to other much more complex objects such as the pig data set.
The minor difference between the Learned-reco and the Ram-Lak-reco in the real CT data experiment prove that the learned filter is not object dependent and not over-fitted to our training data.
Utilizing discs with varying radii introduces two properties to the training process. First, narrow discs model the Dirac-impulse. Secondly, discs with larger radii exhibit large homogeneous regions. Occurring cupping and dc-shift artifacts in the homogeneous area will lead to strong gradients, which appear due to the wrong discretization. As a consequence, the weights converge towards the Ram-Lak solution, which is nothing else than learning the proper discretization of the continuous Ramp. Using the ideal Ramp less cupping is observed. This renders the dc-shift as the dominant artifact to compensate for.
We expect that augmentation approaches of this method will lead to filters that are invariant / less prone to noise characteristics imposed by CT physics intrinsically. Thus, we believe that noise augmentation will lead to filters similar to the Shepp-Logan filter. Augmentation in this context will have a very similar result as a Wiener filter that is optimal given certain noise properties.
Different approaches to learn the reconstruction filter were published in the past. Floyd \cite{Floyd} successfully learned the discrete version of the ramp filter for SPECT image reconstruction. However, compared to our presented method, their approach learns the filter in spatial domain using a neural network with fully connected layers. As a consequence of the system design, Floyd reported practical issues implementing the huge amount of trainable weights at the time of the publication. Furthermore, no detailed analysis of the discretization properties of the learned filter was performed.
Even though similar approaches to learn the reconstruction filter were published, e.g. \cite{Pelt2014,Floyd}, none of them explicitly uses domain knowledge to design the network topology. In contrast, our derivation of the network topology is based on the continuous analytical problem description. Furthermore, the transition to the discrete filter is intrinsically solved due to the discrete nature of the neural networks.
In our view these observations are interesting to gain a better understanding of deep learning techniques and traditional analytic techniques such as Wiener filtering and discretization theory. To the best of our knowledge, we did not observe such links between analytical signal processing theory and deep learning so far.

\section{Conclusion}
We presented  an approach to learn the discrete optimal reconstruction filter directly from the continuous Ramp filter. We have shown that the learning approach will automatically compensate for the errors inflicted by the discretization in an L2-sense optimal way with respect to our given training data. This is achieved by formulating a cost function to learn the filter in the frequency domain. This enables us to initialize the filter with the ideal Ramp, which can be seen as a very strong pre-training. Furthermore, the formulation can be straightforward transfered to a neural network architecture. Combining the proposed solution with the deep neural network suggested by Würfl et al. enables us to provide a fully trainable data-driven reconstruction deep learning approach. In future work, we want to apply noise models to the training data to learn an optimal discrete filter which is less sensitive to noise.

\label{sec:ref}
\bibliographystyle{IEEEbib}
\bibliography{refs}
\end{document}